\documentclass[submission,copyright,creativecommons]{eptcs}
\providecommand{\event}{ICLP 2024} 

\usepackage{iftex}

\ifpdf
  \usepackage{underscore}         
  \usepackage[T1]{fontenc}        
\else
  \usepackage{breakurl}           
\fi

\usepackage{graphicx}
\usepackage[most]{tcolorbox}
\usepackage[title]{appendix}
\usepackage{multirow,tabularx}
\usepackage{svg}
\usepackage{algorithm}
\usepackage[noend]{algpseudocode}
\usepackage{hyperref}
\usepackage[caption=false]{subfig}
\usepackage[inline]{enumitem}
\usepackage{booktabs}
\usepackage{fancyvrb}
\usepackage{fvextra}
\usepackage{listings}
\newtcolorbox[blend into=figures]{floatingtcolorbox}[2][]{size=fbox,float=htb,capture=minipage,
title={#2},every float=\centering,#1}

\newcounter{example}[section]
\newenvironment{example}[1][]{\refstepcounter{example}\par\medskip
   \noindent \textbf{Example~\theexample. #1} \rmfamily}{\medskip}

\title{LLM+Reasoning+Planning for Supporting Incomplete User Queries in Presence of APIs}
\author{Sudhir Agarwal \quad Anu Sreepathy
\institute{Intuit AI Research, Mountain View, CA, USA}
\email{\{sudhir\_agarwal, anu\_sreepathy\}@intuit.com}
\and
David H. Alonso \quad Prarit Lamba
\institute{Intuit Inc., Mountain View, CA, USA}
\email{\{david\_haroalonso, prarit\_lamba\}@intuit.com}
}
\def\titlerunning{LLM+Reasoning+Planning for Incomplete Queries over APIs}
\def\authorrunning{Agarwal, Sreepathy, Alonso \& Lamba}

\begin{document}

\maketitle

\begin{abstract}
Recent availability of Large Language Models (LLMs) has led to the development of numerous LLM-based approaches aimed at providing natural language interfaces for various end-user tasks. These end-user tasks in turn can typically be accomplished by orchestrating a given set of APIs. In practice, natural language task requests (user queries) are often incomplete, i.e., they may not contain all the information required by the APIs. While LLMs excel at natural language processing (NLP) tasks, they frequently hallucinate on missing information or struggle with orchestrating the APIs. The key idea behind our proposed approach is to leverage logical reasoning and classical AI planning along with an LLM for accurately answering user queries including identification and gathering of any missing information in these queries. Our approach uses an LLM and ASP (Answer Set Programming) solver to translate a user query to a representation in Planning Domain Definition Language (PDDL) via an intermediate representation in ASP. We introduce a special API ``get\_info\_api'' for gathering missing information. We model all the APIs as PDDL actions in a way that supports dataflow between the APIs. Our approach then uses a classical AI planner to generate an orchestration of API calls (including calls to get\_info\_api) to answer the user query. Our evaluation results show that our approach significantly outperforms a pure LLM based approach by achieving over 95\% success rate in most cases on a dataset containing complete and incomplete single goal and multi-goal queries where the multi-goal queries may or may not require dataflow among the APIs.
\end{abstract}

\section{Introduction}
\label{sec:intro}
Customers of large organizations have a variety of questions or requests (collectively known as queries in the following) pertaining to the organization’s domain of operation. Providing relevant and accurate responses to such user queries is critical and requires a thorough analysis of the user’s context, product features, domain knowledge, and organization policies. The user queries may encompass a variety of types - data lookup and aggregation queries, help requests, how-to questions, record update requests or a combination of these types. 

Recently, transformer-based large language models (LLMs) have shown wide success on many natural language understanding and translation tasks, also demonstrating some general database querying~\cite{gao2023texttosql,li2024petsql,DBLP:conf/acl/Wang00L23} and reasoning and planning~\cite{DBLP:conf/corl/IchterBCFHHHIIJ22,huang2022inner,liu2023llmp,DBLP:conf/iclr/ZengAICWWTPRSLV23} capability on diverse tasks without having to be retrained. However, the data and knowledge required for accurately answering customers' queries are partly or completely organization internal and not available to LLMs trained on publicly available data. Even in case of organization internal LLM deployments, it is often not feasible to give LLMs direct access to databases for various security and privacy reasons. In lieu of that, organizations develop APIs to make these internal artifacts programmatically accessible to the organization’s applications.

Several frameworks and techniques have been proposed for answering user queries using a combination of LLM and tools/APIs, e.g. LangChain~\cite{langchain}, Gorilla~\cite{patil2023gorilla}, ToolFormer~\cite{schick2023toolformer}, and TravelPlanner~\cite{DBLP:conf/nips/LuPCGCWZG23,xie2024travelplanner}. However, such frameworks rely on LLMs for selecting and composing tools and as a result either do not scale well beyond a small set of APIs/tools or have limited planning and API orchestration capability. These weaknesses limit the use of such frameworks for practical industrial applications. 

To address these limitations, some recent works have investigated the use of an external classical planner along with an LLM. Given a description of the possible initial states of the world, a description of the desired goals, and a description of a set of possible actions, the classical planning problem involves synthesizing a plan that, when applied to any initial state, generates a state which contains the desired goals (goal state)~\cite{DBLP:books/daglib/0014222}. The approaches presented in~\cite{agarwal2024tic,liu2023llmp} have demonstrated that utilizing an LLM to create the task PDDL (a representation of a user query as a planning problem in Planning Domain Definition Language) from a natural language planning task description, and then utilizing an external classical planner to compute a plan, yields better performance than relying solely on an LLM for end-to-end planning. However, these approaches have been shown to support only classical planning tasks, which hinders their use for answering user queries in the presence of APIs. 

Furthermore, all the above mentioned approaches assume complete user queries, i.e., queries that contain all the required information for computing an answer to the query. In practice however, user queries are often incomplete. In general, detecting and gathering missing information depends on the granularity of the underlying atomic actions or APIs as well as dataflow among them at runtime. For example, if a user wants to book a flight and provides the source and destination airports information but the flight booking API requires the travel date as well, the user query is considered incomplete with respect to the available APIs. The AutoConcierge framework~\cite{DBLP:conf/padl/ZengRPBAG24} can detect missing information for a pre-defined goal assuming that the required information for accomplishing the goal is known a-priori. However, there is still a need for an approach that can handle different kinds of possibly incomplete queries.


\begin{figure}[ht]
  \centering
  \includegraphics[width=\linewidth]{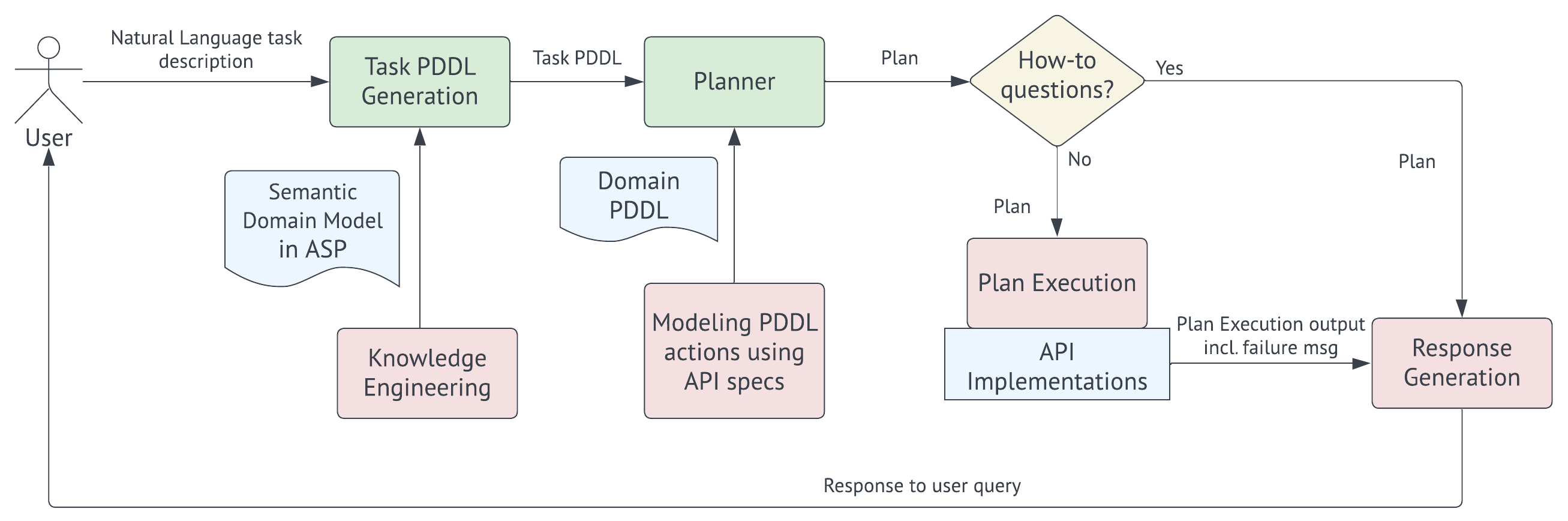}
  \caption{Overview of user query answering using LLMs and Classical Planning }
  \label{fig:e2e-Diagram}
\end{figure}

Figure~\ref{fig:e2e-Diagram} presents the high level architecture of our approach for supporting several kinds of user queries using a given set of APIs. We translate a user query to a task PDDL (query's representation in PDDL) and use a classical AI planner for orchestrating APIs (plan) for the generated task PDDL. The plan execution component executes the plan by invoking the APIs in the specified order. For how-to questions, the plan is not executed but sent to the response generation component. Finally, the response generation component generates the overall response to be sent to the user from the individual outputs of the API calls. 
In this paper, our focus is on the Task PDDL Generation and Planner components in particular for supporting incomplete user queries. 

\begin{figure}[ht]
  \centering
  \includegraphics[width=\linewidth]{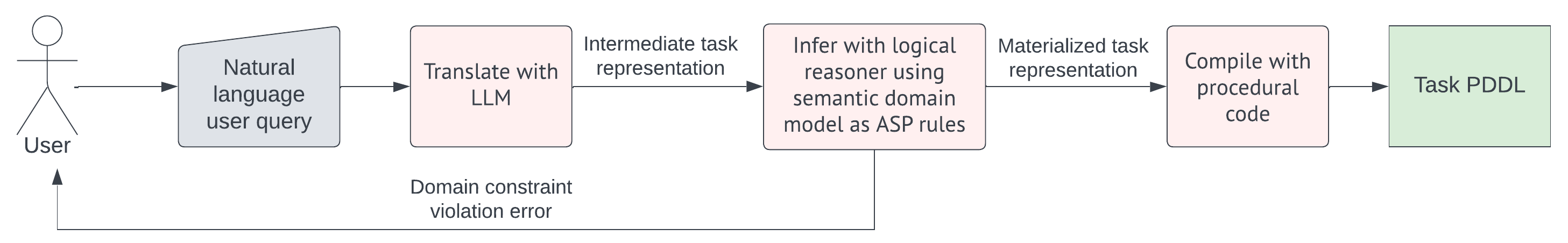}
  \caption{Steps for translating a user query to task PDDL}
  \label{fig:Query-2-Task-PDDL}
\end{figure}

Figure~\ref{fig:Query-2-Task-PDDL} illustrates our process of translating a user query to task PDDL by using a novel combination of an LLM and logical reasoning using Answer Set Programming (ASP)~\cite{DBLP:journals/cacm/BrewkaET11,DBLP:conf/aaai/Lifschitz08,DBLP:conf/iclp/GelfondL88}. We use an LLM to generate an intermediate representation of a user query in ASP. Our LLM prompting technique is generic and allows a set of possible user goal specifications to be plugged in. This step is described in Section~\ref{sec:query-ASP-representation}. Such intermediate representations allow us to use an ASP solver to deterministically infer additional information, detect inconsistencies in user queries with respect to domain constraints, and bridge the syntactic and semantic heterogeneities between a user query and the target task PDDL. We refer to the union of facts in the intermediate representation and the inferred information as materialized representation of the user query. This step is described in Section~\ref{sec:int-rep-to-mat-rep}. In cases, where an intermediate representation violates any domain constraints, the materialized representation contains corresponding errors. In these cases, we send the errors back to user. In other cases, we obtain the task PDDL by converting the materialized representation which is in the ASP syntax to PDDL syntax using deterministic procedural code. This step is described in Section~\ref{sec:task-pddl-generation}.


In the next step, we use a classical planner with the task PDDL and an offline created PDDL domain model which includes domain concepts as predicates and specification of the APIs as PDDL actions in terms of these domain predicates (Section~\ref{sec:API Specification}).
In addition to the given set of functionality providing APIs, we introduce a special API \texttt{get\_info\_api} for gathering missing information from the user or an external system at runtime in order to support incomplete queries. The planner returns a plan (including calls to get\_info\_api in case of incomplete queries) such that the execution of the plan computes the answer to the user query. The plan generation step is described in Section~\ref{sec:plan-generation}.

Since there aren't any benchmark datasets of incomplete queries to be answered using APIs, we generated a dataset containing single goal and multi-goal complete and incomplete natural language queries based off a set of APIs described in Section~\ref{sec:API Specification}. We refer to a domain concept in a user query as a goal. Our evaluation results on this dataset show that our approach significantly outperforms a pure LLM based approach by achieving over 95\% success rate in most cases.

\section{Specification of APIs as PDDL Actions}
\label{sec:API Specification}
Throughout this paper, we use the following APIs which are derived from the set of publicly available Intuit Developer APIs\footnote{\url{https://developer.intuit.com/app/developer/homepage}} for experimental purposes.
\begin{itemize*}
    \item \textit{Profit and loss report API}: Generates profit and loss report for a given time period.
    \item \textit{Expense and spend report API}: Generates expense and spend report for a given time period.
    \item \textit{Invoices and sales report API}: Generates invoices and sales report for a given time period.
    \item \textit{Charge lookup API}: Generates detailed report for a given charge amount on a given date.
    \item \textit{Help API}: Provides answer to a given how-to question in a product.
    \item \textit{Contact API}: Connects customer to a human customer agent over a given communication channel for a conversation on a given topic
    \item \textit{Advice API}: Provides advice for a given personal finance or a small business relation question.
    \item \textit{Create invoice API}: Creates a new invoice for given amount and invoice detail.
    \item \textit{Update customer API}: Updates a customer profile with new first name, last name, phone, and email.
\end{itemize*}

In order to be able to use a classical planner for computing an orchestration of available APIs, we model each available API as an action in PDDL. PDDL serves as a standardized encoding of classical planning problems~\cite{Ghallab98,DBLP:series/synthesis/2019Haslum}. A PDDL representation of an action consists of the action's pre-conditions and  effects defined using logical formulas with domain predicates, local variables (action's parameters) and constants. Note that unlike familiar procedural programming languages, PDDL actions' outputs are also declared as part of action's parameters. The PDDL representation of a planning problem is typically separated into two files: a domain PDDL file and a task PDDL file, both of which become inputs to the planner. Broadly, the domain PDDL file includes declaration of  object types, predicates, and specification of actions. 
The task PDDL file provides a list of objects to ground the domain, and the problem’s initial state and goal conditions defined in terms of the predicates. 

Below the PDDL representation of the profit\&loss API as action \texttt{profit\_loss\_api}. The action generates a profit and loss report for given time period. The pre-condition of the action means that variables \texttt{?in1} and \texttt{?in2} have type \texttt{date} as well as have a value (i.e., they are not NULL). The \texttt{?out} var represents the generated report. The pre-condition also includes that the \texttt{?out} must have the type \texttt{profit\_loss\_report} but must not have a value (indicating that \texttt{?out} doesn't represent an already previously generated report). The effects of the action mean that after execution of the action the value of \texttt{?out} is set. Furthermore, the effects mean that after the execution of the action, the generated report \texttt{?out} has \texttt{?in1} and \texttt{?in2} as start date and end date of the generated report \texttt{?out} respectively.

\begin{verbatim}
(:action profit_loss_api
    :parameters (?in1 - var ?in2 - var ?out - var)
    :precondition (and (has_type ?in1 date) (has_value ?in1) 
       (has_type ?in2 date) (has_value ?in2) 
       (has_type ?out profit_loss_report) (not (has_value ?out)))
    :effect (and (start_date ?out ?in1) 
       (end_date ?out ?in2) (has_value ?out)))
\end{verbatim}

A classical planner will find the above action for a user goal requesting a profit and loss report for given start and end dates. However, if the start date or the end date or both are not provided, a planner will fail to find \texttt{profit\_loss\_api} as relevant action. 

We address this problem by introducing a special action \texttt{get\_info\_api} to gather information from the user or an external system at runtime. We model \texttt{get\_info\_api} as a PDDL action as shown below. The \texttt{get\_info\_api} action requires a variable of a type that is not set and ensures that it is set after the execution of \texttt{get\_info\_api}. 

\begin{verbatim}
(:action get_info_api
    :parameters (?in_var - var ?in_type - var_type)
    :precondition (and (has_type ?in_var ?in_type) 
        (not (has_value ?in_var)))
    :effect (and (has_value ?in_var)))    
\end{verbatim}

\begin{figure}[ht]
     \centering
     \subfloat[Plan for query \textit{Show me my profit and loss report}\label{fig:example1Plan}]{%
        \includegraphics[width=0.35\linewidth]{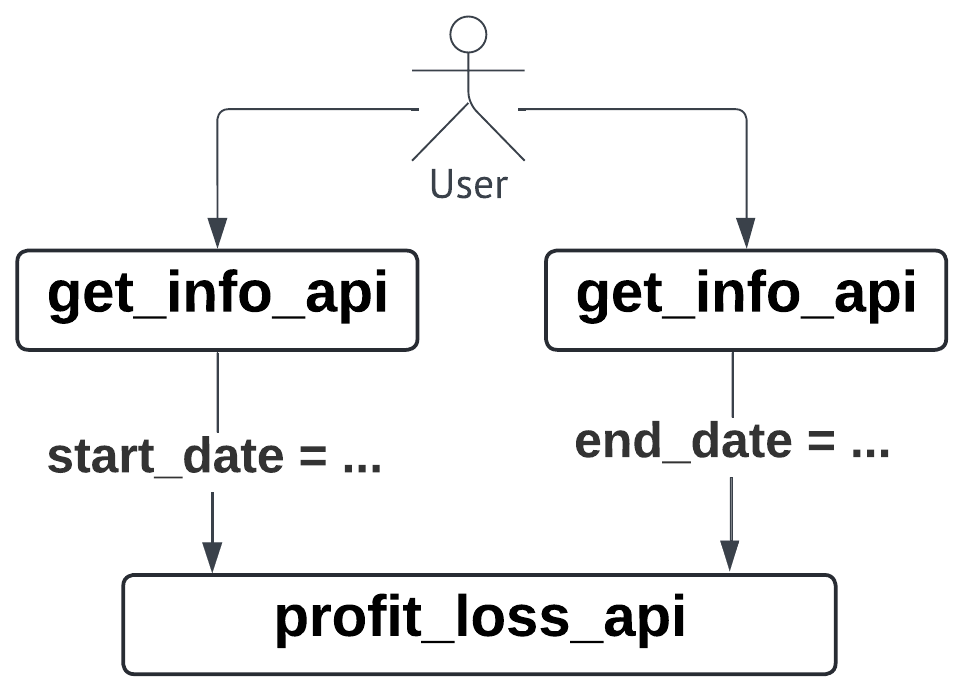}
    }
    \hfill
    \subfloat[Plan for query \textit{Can I see my profit and loss statement from March to May 2023? I would like to discuss my profits further over chat}\label{fig:example2Plan}]{%
        \includegraphics[width=0.6\linewidth]{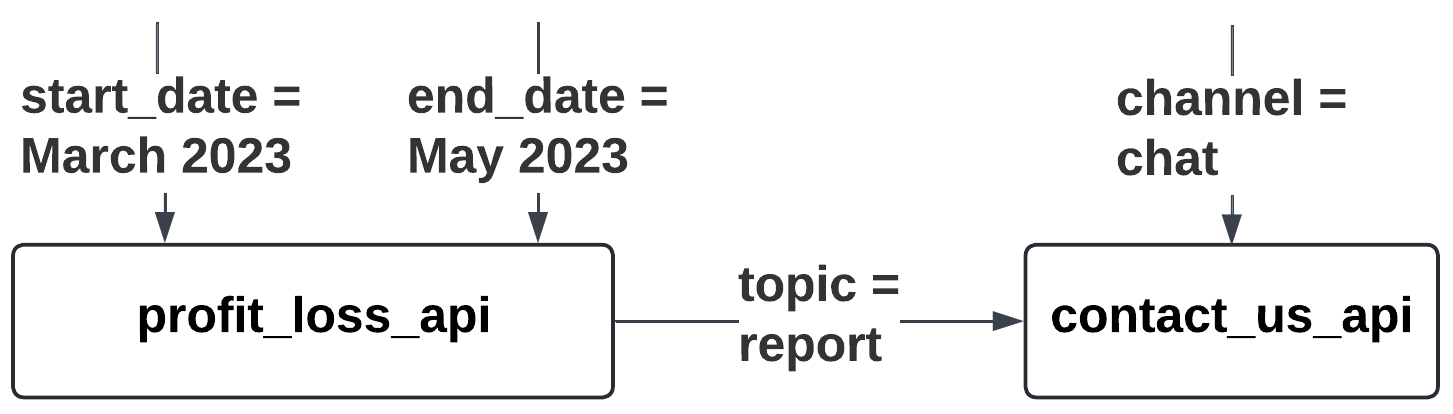}
    }
    \caption{Plan for an incomplete and a complete query}
    \label{fig:three graphs}
\end{figure}

This modeling of \texttt{get\_info\_api} enables a planner to include get\_info\_api calls in the plan for gathering missing information. For example, for the query in Figure~\ref{fig:example1Plan} we aim at detecting the profit \& loss report API, and asking the user for the missing report time period. Similarly, in case of a more complex user query in Figure~\ref{fig:example2Plan}, we aim at detecting the profit \& loss report API and the contact API as well as the profit \& loss report as the conversation topic with the customer agent. 

For the purpose of this paper, we have modeled the domain PDDL manually. Efficient authoring of domain PDDL is out of scope of this work. However, we would like to point that approaches such as~\cite{DBLP:conf/nips/GuanVSK23} may be leveraged for (semi-) automatically generating the domain PDDL for large domains. Refer to Appendix~\ref{appendix:domain-pddl} for the specification of all APIs in our dataset.

\section{User Query to ASP Representation}


As illustrated in Figure~\ref{fig:Query-2-Task-PDDL}, in order to generate task PDDL for a user query, in the first step, we use an LLM for translating the user query to an intermediate representation in ASP. The main reason behind this step is that LLMs perform well on such translation tasks while they hallucinate when they are also required to generate logically derivable information~\cite{DBLP:conf/nips/ZelikmanHPGH23,DBLP:conf/aiide/KellyCWM23,DBLP:journals/mima/FloridiC20,DBLP:conf/nips/Wei0SBIXCLZ22}. In the second step, we use a logical reasoner for inferring other information similar to approaches presented in~\cite{DBLP:journals/corr/abs-2302-03780,DBLP:conf/acl/YangI023,agarwal2024tic}.

\subsection{User Query to Intermediate Representation}
\label{sec:query-ASP-representation}
We construct the LLM prompt with the following steps for translating user query to an intermediate representation in ASP.

\paragraph{Step 1: Define a set of supported goals.} The set of goals doesn't need to have 1:1 correspondence with the set of APIs. But, the set of goals corresponds to expected user requests. Such a modeling enables decoupling of user requests from APIs as the end users can not be expected to be familiar with the APIs (cf. OpenAI function calling approach~\footnote{\url{https://platform.openai.com/docs/guides/function-calling}}).
\paragraph{Step 2: Describe argument types.} For each argument of the supported goals, define the type by giving a few examples or the set of possible values as appropriate. Below example defines argument types for date period and communication channel. See Appendix~\ref{appendix:argument-types} for definition of all argument types for our dataset.
\begin{tcolorbox}[size=fbox,top=0pt,bottom=0pt,left=0pt,right=0pt,enhanced,breakable]
\small
\begin{verbatim}
arg_type_date_period = {"examples": {"nov 2023": ("11/01/2023",
"11/30/2023"), "fy21": ("01/01/2021", "12/31/2021"), ...}

arg_type_comm_channel = {"possible_values": ["video", "chat", "phone"]}
\end{verbatim}
\end{tcolorbox}

\paragraph{Step 3: Describe domain goals.} Describe each goal using a name, description and required information for the goal. Refer to Appendix~\ref{appendix:domain-goals} for complete list of supported domain goals.
\begin{tcolorbox}[size=fbox,top=0pt,bottom=0pt,left=0pt,right=0pt,enhanced,breakable]
\small
\begin{verbatim}
{"name": "goal_1", "description": "request for report on profit, loss, 
earnings, business insights, revenue, figures.", "required information":
[{"name": "report_period", "description": "time period of the requested 
report defined by start and end dates.", "type" : arg_type_date_period}]}
\end{verbatim}
\end{tcolorbox}

\paragraph{Step 4: Define instructions.} We instruct the LLM to extract goals and required information from the user query. 
\begin{tcolorbox}[size=fbox,top=0pt,bottom=0pt,left=0pt,right=0pt,enhanced,breakable]
\small
\begin{verbatim}
Given goal types with their required information. Extract from the 
provided user query: 
1. The one or more goals of the query from the given set of goals.
Represent each extracted goal <x> of type <T> as "_goal(<x>, <T>).".
2. If the user query contains any required information for the extracted
goal, then extract that too. While doing so, if possible values 
are defined for the argument, then choose one from them if applicable.
\end{verbatim}
\end{tcolorbox}

\paragraph{Step 5: Construct LLM prompt.} LLM prompt also includes a few in-context examples that are independent of the domain of our dataset. Refer to Appendix~\ref{appendix:in-context-examples} for complete list of in-context examples. 
\begin{tcolorbox}[size=fbox,top=0pt,bottom=0pt,left=0pt,right=0pt,enhanced,breakable]
\small
\begin{verbatim}
<Instructions as described above>

Below a few examples of goals, text and the answer.
\end{verbatim}
\texttt{<As in Appendix~\ref{appendix:in-context-examples}>}
\begin{verbatim}
Goals: """ <Domain goals as described above.> """

Text: """ <user query> """

Answer:
\end{verbatim}
\end{tcolorbox}

Below are a few example queries and their respective intermediate representations in ASP as returned by the LLM.


\noindent
\begin{minipage}{.48\textwidth}
\begin{example}
\label{ex:query1}
\textit{Show me 2023 Q1 detailed expense report}.
\label{exm:int-rep-1}
\vspace{-7pt}
\begin{verbatim}
_goal(x, goal_2).
_report_period(x, ("01/01/2023",
    "03/31/2023")).    
\end{verbatim}
\end{example}
\vspace{-7pt}
\begin{example}
\label{ex:query2}
\textit{Provide me with the profit and loss statement for the previous quarter and put me on a phone call with a representative to discuss it}.
\label{exm:int-rep-2}
\vspace{-7pt}
\begin{verbatim}
_goal(x, goal_1).
_report_period(x, ("07/01/2024",
    "09/30/2024")).
_goal(y, goal_4).
_contact_topic(y, x).
_contact_channel(y, "phone").
\end{verbatim}
\end{example}
\end{minipage}
\hfill
\begin{minipage}{.48\textwidth}
\vspace{-45pt}
\begin{example}
\label{ex:query4}
\textit{Profit and loss report}.
\label{exm:int-rep-4}
\vspace{-7pt}
\begin{verbatim}
_goal(x, goal_1).
\end{verbatim}
\vspace{-7pt}
\end{example}
\begin{example}
\label{ex:query3}
\textit{I want to chat with a representative}.
\label{exm:int-rep-3}
\vspace{-7pt}
\begin{verbatim}
_goal(x, goal_6).
_contact_channel(x, "chat").
\end{verbatim}
\end{example}
\vspace{-7pt}
\begin{example}
\label{ex:query5}
\textit{Show me expense report from July 2024 to Jan 2024}.
\label{exm:int-rep-5}
\vspace{-7pt}
\begin{verbatim}
_goal(x, goal_2).
_report_period(x, ("07/01/2024",
    "01/31/2024")).    
\end{verbatim}
\end{example}
\end{minipage}

The query in Example~\ref{ex:query1} is a complete query. The query in Example~\ref{ex:query2} is a complete query with two goals and dataflow. The profit \& loss report \texttt{x} is the topic of the conversation for the contact \texttt{y}. The queries in Example~\ref{ex:query4} and Example~\ref{ex:query3} are incomplete queries as the query in Example~\ref{ex:query4} doesn't contain start and end dates of the report and the query in Example~\ref{ex:query3} doesn't contain the conversation topic. The query in Example~\ref{ex:query5} contains both the start date and the end date but violates the domain constraint that the end date must be after the start date.

\subsection{Intermediate Representation to Materialized Representation}
\label{sec:int-rep-to-mat-rep}
An intermediate representation captures the content of the user query using formats and predicates that are closer to those of typical user utterances. In general, user queries cannot be expected to be formulated using the same vocabulary and format as the arguments of the APIs. In this step, we infer additional information as well as bridge the syntactic and semantic gaps. We accomplish this by using an ASP solver, with the intermediate representation and domain rules as inputs. For our current implementation we use Clingo~\cite{DBLP:journals/tplp/GebserKKS19} python package\footnote{\url{https://pypi.org/project/clingo/}} as the ASP solver.

Below a snippet of the domain rules for our dataset (see Appendix~\ref{appendix:domain-rules} for all domain rules). Note that even though the domain rules needed for our current dataset are rather simple and few in number, our framework of first translating the query to an intermediate representation in ASP allows us to plug-in a large number of complex rules if needed.
\begin{tcolorbox}[size=fbox,top=0pt,bottom=0pt,left=0pt,right=0pt,enhanced,breakable]
\small
\begin{verbatim}
goal(X, profit_loss_report) :- _goal(X, goal_1).
start_date(X, Y, date) :- goal(X, profit_loss_report), _report_period(X, (Y,_)).
end_date(X, Y, date) :- goal(X, profit_loss_report), _report_period(X, (_,Y)).

goal(X, contact_us) :- _goal(X, goal_4).
contact_topic(X, Y, string) :- goal(X, contact_us), _contact_topic(X, Y).
contact_channel(X, Y, string) :- goal(X, contact_us), _contact_channel(X, Y).
...
error("end date must be after start date") :- start_date(X, D1, date),
    end_date(X, D2, date), false == @lte_dates(D1, D2).
\end{verbatim}
\end{tcolorbox}

The first rule translates the goal type to the type used in the vocabulary of the domain PDDL. The second and third rules infer \texttt{start\_date} and \texttt{end\_date} from the the user provided report\_period. These rules also add the data types \texttt{date}, \texttt{string} for the values to facilitate the planning in the later step. The last rule infers an error when the end date is before the start date. In general, this technique allows us to generate error messages for complex constraint violations using ASP. For our example queries in Section~\ref{sec:query-ASP-representation}, the ASP solver returns below materialized representations after applying the domain rules on the intermediate representations of the queries.

\noindent
\begin{minipage}{.5\textwidth}
Materialized representation for Example~\ref{exm:int-rep-1}:\vspace{-7pt}
\begin{verbatim}
goal(x, expense_spend_report).
start_date(x, "01/01/2023", date).
end_date(x, "03/31/2023", date).
\end{verbatim}
Materialized representation for Example~\ref{exm:int-rep-2}:\vspace{-7pt}
\begin{verbatim}
goal(x, profit_loss_report).
start_date(x, "07/01/2024", date).
end_date(x, "09/30/2024", date).
goal(y, contact_us).
contact_topic(y, x, string). 
contact_channel(y, "phone", string).
\end{verbatim}
\end{minipage}
\begin{minipage}{.5\textwidth}
\vspace{8pt}
Materialized representation for Example~\ref{exm:int-rep-4}:\vspace{-7pt}
\begin{verbatim}
goal(x, profit_loss_report).
\end{verbatim}
Materialized representation for Example~\ref{exm:int-rep-3}:\vspace{-7pt}
\begin{verbatim}
goal(x, contact_us).
contact_channel(x, "chat", string).
\end{verbatim}
Materialized representation for Example~\ref{exm:int-rep-5}:\vspace{-7pt}
\begin{verbatim}
goal(x, expense_spend_report).
start_date(x, "07/01/2024", date).
end_date(x, "01/31/2024", date).
error("start date is after end date.").
\end{verbatim}
\vspace{15pt}
\end{minipage}
Note that the materialized representation of Example~\ref{exm:int-rep-5} contains an error atom because the end date is before the start date. In such cases, we do not continue with task PDDL generation and send the error back to the user (see also Figure~\ref{fig:Query-2-Task-PDDL}). 

\section{Orchestrate APIs using Planner}
\label{sec:planning}

\subsection{Task PDDL Generation}
\label{sec:task-pddl-generation}
\begin{figure}[ht]
  \centering
  \includegraphics[width=\linewidth]{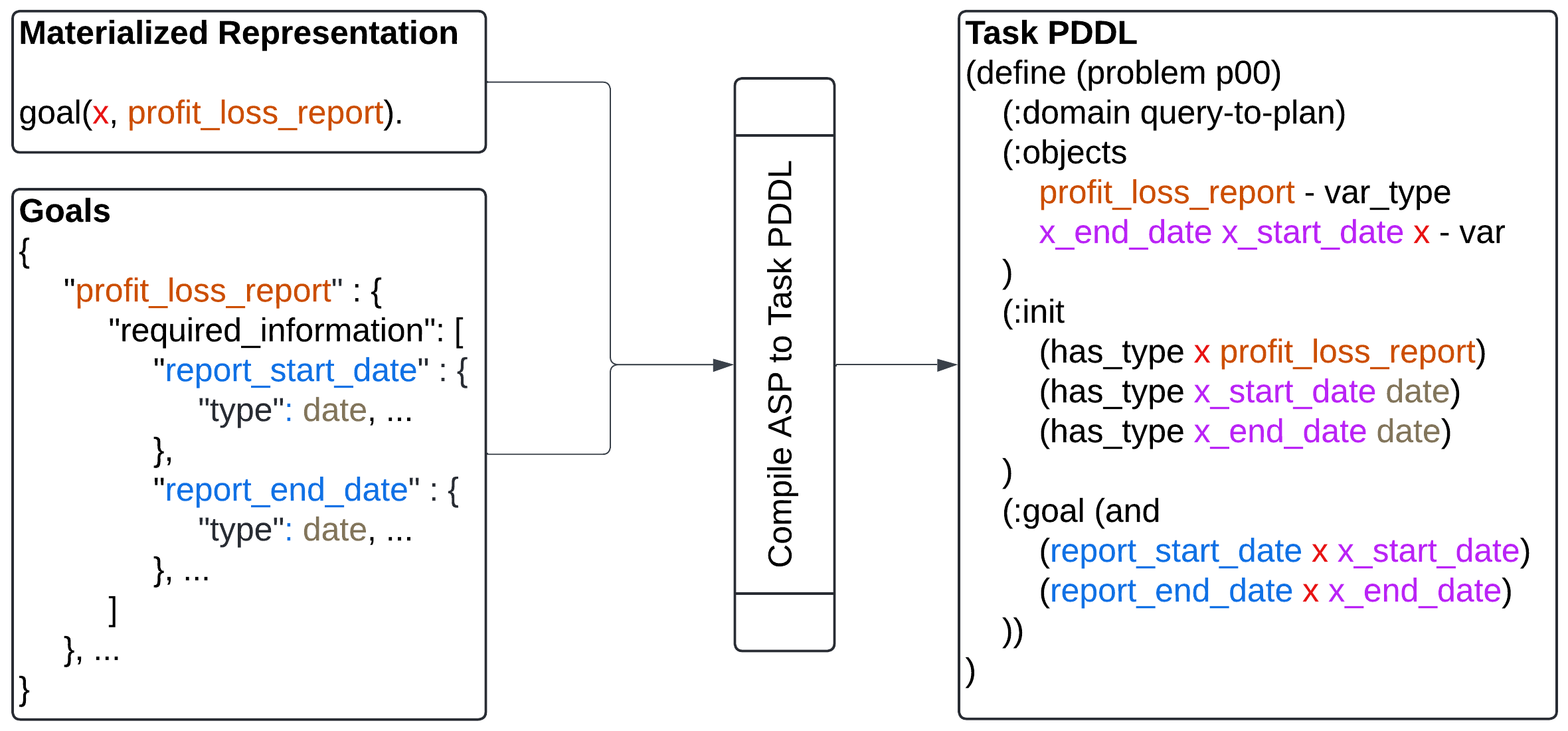}
  \caption{Materialized representation to task PDDL for the  query \textit{Profit and loss report}.}
  \label{fig:ASPToTaskPDDL}
\end{figure}

A materialized representation contains all user provided information in the target terminology and format. The next and the last step is to generate a plan. In order to be able to do that, we need to convert the materialized representation to a PDDL representation (task PDDL).

Figure~\ref{fig:ASPToTaskPDDL} illustrates this process using Example~\ref{ex:query1}. Every goal $x$ becomes a $var$ and every goal type $t$ becomes a $var\_type$. For each goal $x$ of type $t$, (a) add $(has\_type \; x \; t)$ to the init section, (b) for each argument $a$ of $t$ and predicate $p$, a var $x\_a$ is added to the objects, $(has\_type \; x\_a \; t)$ is added to init, $(p \; x \; x\_a)$ is added to goal, and if $x\_a$ has a value $v$, then $(has\_value \; x\_a \; v)$ is added to init. Refer to Appendix~\ref{appendix:mat-rep-to-task-pddl-algo} for the complete algorithm for generating materialized representation to task PDDL. The output of the algorithm, the task PDDL for Example~\ref{ex:query1} is shown on the right side Figure~\ref{fig:ASPToTaskPDDL}. Refer to Appendix~\ref{appendix:example-task-pddl} for the task PDDLs of other example queries.

\subsection{Plan generation}
\label{sec:plan-generation}
Once the task PDDL is generated, all we need to do is to call a PDDL planner with the task PDDL and the domain PDDL. In our implementation we use the Fast Downward Planner~\footnote{\url{https://www.fast-downward.org/HomePage}}~\cite{Helmert2006} with configuration parameters \textit{alias = lama} and \textit{search-time-limit = 1}. In other implementations, where compatibility to PDDL may not be important, one may also choose an appropriate ASP based planner~\cite{DBLP:journals/tplp/SonPBS23}. 

Using an external classical AI planner has several benefits such as:
\begin{itemize*}
    \item \textit{Scalability}: AI planners scale well wrt number of APIs as long as the functionality of APIs can be defined in terms of (Inputs, Outputs, Preconditions, Effects) with logical formulas.
    \item \textit{Support for interaction}: In case of incomplete queries the generated plan includes calls to get\_info\_api API for gathering information from user
    \item \textit{Optimality}: APIs can be assigned a cost; Planner computes an optimal plan wrt the cost function.
    \item \textit{Graceful failure}: For out of domain queries planner won’t generate a plan rather than hallucinating.
\end{itemize*}

For the example query \textit{Show me 2023 Q1 detailed expense report}, the planner generates the plan:
\begin{verbatim}
Step 1. x_start_date = "01/01/2023";
Step 2. x_end_date = "03/31/2023";
Step 3. x = expense_spend_api(x_start_date, x_end_date);
\end{verbatim}

For the example query \textit{Provide me with the profit and loss statement for the previous quarter and then put me on a phone call with a representative to discuss it}, the planner generates:
\begin{verbatim}
Step 1. x_start_date = "07/01/2024";
Step 2. x_end_date = "09/30/2024";
Step 3. y_contact_channel = "phone";
Step 4. x = profit_loss_api(x_start_date, x_end_date);
Step 5. y = contact_us_api(x, y_contact_channel);
\end{verbatim}
Note that the the contact topic is bound to the generated profit and loss report \texttt{x}. 

For the example query \textit{I want to chat with a representative}, the planner generates:
\begin{verbatim}
Step 1. x_contact_topic = get_info_api("contact topic", date);
Step 2. x_contact_channel = "chat";
Step 3. x = contact_us_api(x_contact_topic, x_contact_channel);
\end{verbatim}

For the example query \textit{Profit and loss report}, the planner generates: 
\begin{verbatim}
Step 1. x_start_date = get_info_api("start date", date);
Step 2. x_end_date = get_info_api("end date", date);
Step 3. x = profit_loss_api(x_start_date, x_end_date);
\end{verbatim}

\section{Experiments}
In this section we present the evaluation results of our approach on a generated dataset containing natural language user queries related to various topics such as generation of profit \& loss reports, invoice creation, and how-to help requests.

\subsection{Dataset Generation} 
The initial step in the dataset generation process involves using GPT-4 to generate user queries that represent single goal tasks executable via a subset of the APIs described in Section~\ref{sec:API Specification}. GPT-4 is prompted with instructions and in-context examples to guide the generation process and ensure that the resulting queries align with the requirements of the API. Refer to Appendix~\ref{appendix:dataset-generation-prompt} for an example LLM prompt for dataset generation.

We use the same process to create more complex multi-goal queries simulating a real-world scenario where a user might seek to perform a series of actions in a single request. For example, ``\textit{Can I see my profit and loss statement from March to May 2023? I would like to discuss my profits further over chat.}''. GPT-4 is prompted to generate coherent sequences where the output of one goal execution would become the input of another (multi-goal with dataflow), and complex queries which required multiple APIs to be executed independently (multi-goal without dataflow). 

Once a sufficient number of single and multi-goal queries are generated, we first manually select queries that are representative of real user queries. Then, we manually annotate the selected queries with the ground truth values for the APIs and entities as their arguments. Refer to Appendix~\ref{appendix:dataset-samples} for some sample data in the dataset.



\subsection{Results and Analysis} 
We consider a query as successfully processed iff the generated plan for answering the query contains all the ground truth APIs with correct entities as their arguments. In particular, the \texttt{get\_info\_api} calls correspond to missing entity values in incomplete queries.
This allows us to also measure the success rate of incomplete queries where the planner should generate \texttt{get\_info\_api} actions for missing entities instead of the LLM hallucinating on entity values not present in the query. In our evaluation, a processed query is either correct or wrong, and never fractionally correct. 

\begin{table}[t]
\centering
\caption{Success rate \% of our approach compared with a baseline of end-to-end LLM based approach on single goal queries}
\label{tbl:total-single-plugin-queries}
\begin{tabularx}{\textwidth} { 
  m{1.2in}
  m{0.2in}
  >{\centering\arraybackslash}X
  >{\centering\arraybackslash}X
  >{\centering\arraybackslash}X
  >{\centering\arraybackslash}X
  m{0.2in}
  >{\centering\arraybackslash}X
  >{\centering\arraybackslash}X
  >{\centering\arraybackslash}X
  >{\centering\arraybackslash}X}
 \toprule
 & \multicolumn{5}{c}{\textbf{Complete Queries}} & \multicolumn{5}{c}{\textbf{Incomplete Queries}} \\ 
 & \textbf{\#} & \multicolumn{2}{c}{\textbf{GPT-4}} & \multicolumn{2}{c}{\textbf{GPT3.5}} & \textbf{\#} & \multicolumn{2}{c}{\textbf{GPT4}} & \multicolumn{2}{c}{\textbf{GPT3.5}} \\ 
 & & Base-line & Our Approach & Base-line & Our Approach & & Base-line & Our Approach & Base-line & Our Approach\\
 \midrule
 profit \& loss report & 70 & 22.86 & \textbf{98.57} & 81.43 & \textbf{100} & 2 & 0 & \textbf{100} & 0 & \textbf{100}\\
 \midrule
 expense report & 42 & 23.81 & \textbf{100} & 90.48 & \textbf{100} & 0 & - & - & - & -\\
 \midrule
 invoice sales report & 33 & 54.55 & \textbf{90.91} & 84.85 & \textbf{93.94} & 12 & 0 & \textbf{100} & 0 & \textbf{91.67}\\
 \midrule
 charge lookup & 33 & 81.82 & \textbf{100} & \textbf{96.97} & 93.94 & 5 & 40.00 & \textbf{100} & 0  & \textbf{100}\\
\midrule
 how-to help & 60 & 68.33 & \textbf{98.33} &68.33 & \textbf{90.00} & 0 & - & - & - & -\\
\midrule
 contact us request & 10 & 40.00  & \textbf{100} & 70.00 & \textbf{100} & 47 & 0 & \textbf{91.49} & 0 & \textbf{85.11}\\
\midrule
 financial advice & 100 & 81.00 & \textbf{94.00} & 94.00  & \textbf{97.00} & 0 & - & - & - & -\\
\midrule
 create invoice & 40 & 57.50 & \textbf{100} & \textbf{100} & \textbf{100} & 20 & 0 & \textbf{100} & 0 & \textbf{100}\\
\midrule
update customer & 3 & 0 & \textbf{100} & \textbf{100} & \textbf{100} & 30 & 6.67 & \textbf{100} & 6.67 & \textbf{100}\\
\bottomrule
\end{tabularx}
\end{table}

\begin{table}[t]
\centering
\caption{Success rate \% of our approach compared with a baseline of end-to-end LLM based approach on multi-goal queries.}
\label{tbl:total-multi-plugin-queries}
\begin{tabularx}{\textwidth} { 
  m{1.3in}
  m{0.2in}
  >{\centering\arraybackslash}X
  >{\centering\arraybackslash}X
  >{\centering\arraybackslash}X
  >{\centering\arraybackslash}X
  m{0.2in}
  >{\centering\arraybackslash}X
  >{\centering\arraybackslash}X
  >{\centering\arraybackslash}X
  >{\centering\arraybackslash}X}
  \toprule
 & \multicolumn{5}{c}{\textbf{Complete Queries}} & \multicolumn{5}{c}{\textbf{Incomplete Queries}} \\
 & \textbf{\#} & \multicolumn{2}{c}{\textbf{GPT-4}} & \multicolumn{2}{c}{\textbf{GPT-3.5}} & \textbf{\#} & \multicolumn{2}{c}{\textbf{GPT-4}} & \multicolumn{2}{c}{\textbf{GPT-3.5}} \\
& & Base-line & Our Approach & Base-line & Our Approach & & Base-line & Our Approach & Base-line & Our Approach\\
 \midrule
 2 APIs w/o dataflow & 15 & 0  & \textbf{100} & 0 & \textbf{100} & 10 & 0  & \textbf{100} & 0 & \textbf{100}\\
 \midrule
 2 APIs with dataflow & 20 & 0 & \textbf{90} & 0  & \textbf{70} & 10 & 0 & \textbf{80} & 0 & \textbf{60}\\
 \midrule
 3 APIs with dataflow & 4 & 0 & \textbf{100} & 0 & \textbf{75} & 16 & 0 & \textbf{75} & 0 & \textbf{62.50}\\
 \bottomrule
\end{tabularx}
\end{table}

Table~\ref{tbl:total-single-plugin-queries} and Table~\ref{tbl:total-multi-plugin-queries} present the average success rate (with a variance of 1.0) of our system over five runs on single goal and multi-goal queries respectively. The rows denote the different types of queries in our dataset. Columns 2 and 7 denoted by \# represent the number of complete and incomplete queries respectively. In case of single goal queries, we report success rate for each goal type. In case of multi-goal queries, we distinguish between queries with 2 goals and 3 goals with or without dataflow. A query contains at least one goal and zero or more entities as arguments of the goals. The success rates reported in Table~\ref{tbl:total-single-plugin-queries} and Table~\ref{tbl:total-multi-plugin-queries} are at most equal to the smaller of API orchestration success rate and entity values extraction success rate of the respective classes. See Appendix ~\ref{appendix:evaluation-results} for API orchestration success rates and entity values extraction success rates. 

We compare our method to a baseline where an LLM alone extracts the goals and entities in a query and performs orchestration of APIs. The baseline utilizes function calling method from ~\cite{OpenAIFunctionCalling} where APIs represented as function descriptions are used by the LLM to translate natural language query into function calls. Refer to Appendix~\ref{appendix:baseline-prompt} for the LLM prompt used for the baseline approach. In our experiments, we observe that our approach significantly outperforms the baseline in most cases for single goal queries. For complete queries, the baseline approach often fails to detect the correct goal or extract the entities in a query correctly. The former is mainly due to overlap in the API functionalities and thus the goals, e.g., there are three report generating APIs. The latter is due to large variation in expressing the same entity value. In addition, the baseline approach performs poorly on incomplete queries. In particular, the baseline approach with GPT-4 asks unnecessary clarification questions in case of complete queries and both GPT-4 and GPT-3.5 hallucinate on missing entity values in case of incomplete queries. We also observe that our approach can handle multi-goal complete and incomplete queries with high success rate while the baseline completely fails to orchestrate these queries correctly. 

Overall, the increase in success rate in our approach can be attributed to the use of an LLM only for translating a user query coupled with the use of deterministic tools such as a logical reasoner and a planner for inferring additional information and generating a plan respectively. In particular, using an LLM to translate to an intermediate representation that is closer to the user query increases the translation accuracy as well as minimizes the hallucination. Furthermore, using a logical reasoner facilitates accurate mapping to target schema with the help of ASP rules even in complex domains where an LLM would often generate incorrect inferences. Similarly, using an external planner computes only feasible plans. In case of single goal complete queries, the increase in success rate is due to the use of intermediate representation and reasoning, and the planner doesn't add any additional value as the materialized representation itself can be seen as an equivalent to a plan. In case of single goal or multi-goal incomplete queries as well multi-goal complete queries with dataflow, the increase in the success rate  is due to use of intermediate representation, logical reasoning, and the planner. 

Our approach requires per query one LLM call, one ASP solver call, and one planner call. The total execution time for processing one query in case of GPT-4 is 3--5 seconds and 0.5--1 seconds in case of GPT-3.5. In both cases over 99\% of total time is consumed by the LLM call(s) in the translation step. Note that our LLM response times are measured in a setup with shared resources across all LLM projects within our organization. We believe that the latency will be significantly lower with dedicated LLM access. 

\section{Conclusion}
In this paper, we studied the problem of answering incomplete user queries in presence of APIs. To the best of our knowledge, ours is the first approach to address this problem. Our approach introduces a novel combination of LLMs, logical reasoning, and classical AI planning to support queries that can be complete or incomplete requiring only one API or an orchestration of multiple APIs. Furthermore, our approach supports queries of different kinds such as information seeking queries, how-to queries, and state changing queries. Our evaluation results show that our approach achieves high success rate (over 95\% in most cases including 100\% in some cases). Our approach is generic in the sense that it doesn't depend on a particular set of APIs but allows API specifications to be plugged in. The significant success rate improvement as compared to a pure LLM based baseline can be attributed to the use of interpretable intermediate representation, logical reasoning, and classical AI planning. 

Our approach has a few limitations which we plan to address in our future work. Currently, we send the metadata for all supported goals of the domain to an LLM as part of the prompt. This technique can overshoot the LLM token limit in cases where there are a large number of possible goals in the domain. Currently, our approach only supports queries but not user's soft preferences. One way to address this gap, at least for some types of user preferences, could be to translate them to a cost function which AI planners can directly support. Lastly, the use of AI planner requires the APIs be specified with accurate IOPE specifications in PDDL which may not be applicable for all APIs or difficult to create for APIs with complex functionality.


\bibliographystyle{eptcs}
\bibliography{paper}

\clearpage
\appendix


\section{Translation Prompt}

\subsection{Argument Types}
\label{appendix:argument-types}
\begin{tcolorbox}[left=0pt,right=0pt,enhanced,breakable,size=fbox]
\begin{verbatim}
arg_type_date_period = {'examples': {'nov 2023' : ('11/01/2023',
    '11/30/2023'), 'april 15 2022-june 30 2022' : ('04/15/2022',
    '06/30/2022'), 'fy21': ('01/01/2021','12/31/2021'), '1 Half 2023': 
    ('01/01/2023','06/30/2023'), '6/22 to 7/22': ('06/01/2022',
    '07/31/2022'), '1Q23' : ('01/01/2023', '03/31/2023'), 
    'Mar-Apr 2022': ('03/01/2022','04/30/2022'), 
    'April end-June start 2023': ('04/30/2023', '06/01/2023')}}
    
arg_type_date = {'examples' : {'1 nov 2023' : '11/01/2023', 
    '11th November \'18': '11/11/2018', 'april 15 2022' : '04/15/2022',
    '2/21/18': '02/21/2018'}}

arg_type_amount = {'examples': {'$2': '2.00', '$15.90': '15.99', 
    '$4,500' : '4500.00', '$65': '65.00'}}

arg_type_qb_feature = {'possible_values': ['accounts payable', 
    'add trips manually', 'bank statements', 'budget', 'capital', '
    categorization', 'certification', 'change business name', 
    'connect to bank', 'deposits', 'depreciation',
    'import journal entries', 'inventory', 'melio', 'overtime', 
    'payroll', 'purchase order', 'purchase orders', 'reclassify',
    'recover deleted account', 'reset account', 'record an expense',
    'reconciliation', 'shortcuts', 'timesheets', 'timesheets/payroll', 
    'vendors', 'write off bad debt']}

arg_type_conversation_topic = {'possible_values': [
    'account', 'Accounts Payable', 'Accounts Receivable', 
    'Accounting Software', 'Bank Reconciliation', 'Billing', 
    'Bookkeeping', 'Budget', 'Budget Tracking and Forecasting', 
    'Cash Flow Management', 'Financial Analysis', 'Financial Planning',
    'Financial Reporting', 'Fixed Assets', 'insurance', 
    'Inventory Management', 'Invoicing', 'issue', 'order', 'password',
    'Payroll', 'product', 'Purchase Orders', 'questions', 
    'Reconciliations', 'returns', 'technical', 'shipping', 
    'service_plan', 'tax', 'Tax Filing', 'Vendor Management']}
    
arg_type_conversation_channel = {'possible_values': ['speak', 'talk', 
    'connect', 'video', 'chat', 'phone']}

arg_type_invoice_detail = {'possible_values':['Construction Project',
    'Tutoring Services', 'Website Design', 'Car Repair', 
    'Catering Services', 'Event Management', 'Graphic Design',
    'Photography Service', 'Marketing Campaign', 
    'Business Consultation', 'Furniture Supplies', 'Cleaning Service',
    'Painting Service', 'IT Consultancy', 'Accounting Services',
    'Renovation Work', 'Gardening Service', 'Legal Consultation',
    'Transportation Service', 'Personal Training Services', 
    'catering service', 'marathon coaching', 'construction project',
    'baking class', 'introduction tutorial', 'lawn service',
    'grooming service', 'violin lesson', 'pilates session',
    'IT project', 'yoga class', 'personal training',
    'marketing consultation', 'premium subscription',
    'legal consultation', 'bartending service', 'reiki session',
    'mobile application development', 'home renovation service',
    'furnace inspection', 'dancing lessons', 'car repair service',
    'freelance design work', 'piano lessons', 'cleaning service',
    'plumbing service', 'hairstyling', 'landscaping service',
    'catering service', 'logistics service',
    'personal fitness training', 'graphic design work',
    'babysitting service', 'real estate consultancy', 'SEO services',
    'web development work', 'tailoring service', 'carpentry work',
    'security service', 'digital marketing service']}

arg_type_given_name = {'examples' : ['John', 'Mary']}

arg_type_family_name = {'examples' : ['Smith', 'Fischer']}

arg_type_email = {'examples' : ['j.fischer@abc.com']}

arg_type_phone = {'examples' : ['987-654-3210']}

\end{verbatim}
\end{tcolorbox}

\subsection{Domain Goals}
\label{appendix:domain-goals}
\begin{tcolorbox}[left=0pt,right=2pt,enhanced,breakable,size=fbox]
\begin{verbatim}
[{'type': 'goal_1',
    'description': 'request for generating a report on one of
     [profit_and_loss, income, business_insights, figures, 
     operating_income, report, revenue, earnings].',
    'required information': [{'name': 'report_period',
     'description': 'time period defined by start and end dates. 
      consider leap year while generating feb dates.',
      'type': arg_type_date_period}],
    'examples': [{'Earnings Summary?': '_goal(x, goal_1, earnings).'},
 {'type': 'goal_2',
    'description': 'request for generating a report on one of [expense, 
      spend, bills, operating_expense, spend_figures, spending_insight,
      spend_report, expense_report, business_insights].',
    'required information': [{'name': 'report_period',
      'description': 'time period defined by start and end dates. 
      Consider leap year while generating feb dates.', 'type':
      arg_type_date_period}],
    'examples': [{'What was the total expense for the first quarter of 
    2023?':  '_goal(x, goal_2, expense_report). _report_period(x
      ("01/01/2023","03/31/2023")).'}]},
{'type': 'goal_3',
    'description': 'request for generating a report on one of 
     [earnings, pending_payments, due_accounts, invoices, 
     invoice_report, sales, accrued_expense, financial_forecast, 
     revenue, report].',
    'required information': [{'name': 'report_period',
     'description': 'time period defined by start and end dates. 
     Consider leap year while generating feb dates.', 'type': 
     arg_type_date_period}], 'examples': [{'Show me all the 
     invoices generated between March 1, 2022, and June 1, 2022': 
     '_goal(x, goal_3, invoices).
    _report_period(x, ("03/01/2022","06/01/2022")).'}]},
{'type': 'goal_4', 'description': 'request for one of [charge_lookup]',
    'required_information': [{'name': 'date_of_charge', 'description': 
    'date of charge.', 'type': arg_type_date},{'name': 
    'amount_of_charge', 'description': 'amount of charge.', 'type':
    arg_type_amount}],
    'examples': [{'Why am I being charged $30.00?': '_goal(x, goal_4,
    charge_lookup). _amount_of_charge(x, "30.00").'}]},
{'type': 'goal_5', 'description': 'request for instructions 
    on accomplishing a task in quickbooks.','required_information': [
    {'name': 'help_topic', 'description': 'quickbooks product 
    feature relevant for the task to be accomplished', 'type':
    arg_type_qb_feature}],
    'examples': [{'What is the process for approving and 
    fulfilling purchase orders?': '_goal(x, goal_5).
    _help_topic(x, "purchase orders").'}]},
{'type': 'goal_6',
    'description': 'request for a conversation with a person on the 
    best matching conversation topic using best matching 
    conversation medium.', 'required information': [{'name': 
    'contact_topic', 'description': 'topic of conversation.', 'type':
    arg_type_conversation_topic}, {'name': 'contact_channel', 
    'description': 'explicitly mentioned medium of
    conversation.', 'type': arg_type_conversation_channel},],
    'examples': [ {'I have some questions about billing. Can I chat 
    with an expert about it?': '_goal(x, goal_6, expert). 
    _contact_topic(x, "Billing"). _contact_channel(x, "chat").'},
    {'Can I speak to a representative?': '_goal(x, goal_6, 
    representative). _contact_channel(x, "speak").'},
    {'Can I talk to an expert? What is the best way?': '_goal(x, 
    goal_6, expert). _contact_channel(x, "talk").'},
    {'Can I book a phone call with a call agent?': 
    '_goal(x, goal_6, call_agent). _contact_channel(x, "phone").'},
    {'Could I please speak with someone who can answer my questions?': 
    '_goal(x, goal_6, representative). _contact_channel(x, "speak").
    _contact_topic(x, "questions").'}]},
{'type': 'goal_7',
    'description': 'request for an advice about one of ["business 
    analysis", "business comparison", "business recommendation", 
    "personal finance", "business expense", "profit making"].',
    'required information': [],
      'examples': [{'Any advice for dealing with monthly recurring 
      expenses?': '_goal(x, goal_7).'}, {'How does my liquidity compare 
      to similar businesses?': '_goal(x, goal_7).'}]},
{'type': 'goal_8',
    'description': 'request for creating a new invoice for a given 
    amount and invoice detail.',
    'required information': [{'name': 'invoice_amount', 'description':
    'amount of invoice.', 'type': arg_type_amount},
    {'name': 'invoice_detail', 'description': 'detail of invoice.', 
    'type': arg_type_invoice_detail},],
    'examples' : [{'Invoice needed of $200 for grooming service': 
    '_goal(x, goal_8, new_invoice). _invoice_amount(x, "200.00").
    _invoice_detail(x, "grooming service").'}]},
{'type': 'goal_9', 'description': 'request for updating a customer 
    profile', 'required information': [{'name': 'customer_given_name',
    'description': 'customer given name in customer profile.', 'type':
    arg_type_given_name}, {'name': 'customer_family_name', 
    'description': 'customer family name in customer profile.', 'type':
    arg_type_family_name}, {'name': 'customer_email', 
    'description': 'customer email in customer profile.' ,'type':
    arg_type_email}, {'name': 'customer_phone', 'description': 
    'customer phone in customer profile.', 'type': arg_type_phone}],
    'examples': [{'Can you add a new profile for Henry Davis?': 
    '_goal(x, goal_9, customer_profile). _customer_given_name
    (x, "Henry"). _customer_family_name(x, "Davis").'}]
}
]
\end{verbatim}
\end{tcolorbox}


\subsection{In-context Examples}
\label{appendix:in-context-examples}

\begin{tcolorbox}[left=0pt,right=0pt,enhanced,breakable,size=fbox]
\begin{verbatim}
arg_type_color = {'name': 'color', 'description': 'color of an object',
    'type': {'description': 'color of an object', 'possible_values':
    ['red', 'green', 'blue', 'yellow']}}

arg_type_shape = {'name': 'shape', 'description': 'shape of an object',
    'type': {'description': 'shape on an object', 'possible_values':
    ['large', 'big', 'small', 'medium']}}

ex_goal_types = [
    {'type': 'fruits_goods', 'description': 'request for report about
    one of [apple, orange, ball].', 'required information':
    [arg_type_color, arg_type_shape]},
]
\end{verbatim}
\end{tcolorbox}

\begin{tcolorbox}[left=0pt,right=0pt,enhanced,breakable,size=fbox]
\begin{verbatim}
Goals: <AS ABOVE>

Text: show me red apples.

Answer:
% --- begin ---
_goal(x, fruits_goods, apple).
_color(x, "red").
% --- end ---

Goals: <AS ABOVE>

Text: which large oranges are green.

Answer:
% --- begin ---
_goal(x, fruits_goods, orange).
_color(x, "green").
_shape(x, "large").
% --- end ---

Goals: Goals: <AS ABOVE>

Text: big blue balls.

Answer:
% --- begin ---
_goal(x, fruits_goods, ball).
_color(x, "blue").
_shape(x, "big").
% --- end ---

Goals: Goals: <AS ABOVE>

Text: red oranges

Answer:
% --- begin ---
_goal(x, fruits_goods, orange).
_color(x, "red").
% --- end ---

Goals: Goals: <AS ABOVE>

Text: list of small oranges that are yellow

Answer:
% --- begin ---
_goal(x, fruits_goods, orange).
_color(x, "yellow").
_shape(x, "small").
% --- end ---

\end{verbatim}
\end{tcolorbox}

\section{Domain Modeling}
\subsection{Domain PDDL}
\label{appendix:domain-pddl}
\begin{tcolorbox}[left=0pt,right=0pt,enhanced,breakable,size=fbox]
\begin{verbatim}
(define (domain gen-orch-planner)
    (:requirements :strips)
    (:types
        var - object
        var_type - object
    )
    (:predicates
        (report_start_date ?r - var ?t - var)
        (report_end_date ?r - var ?t - var)
        (charge_date ?r - var ?t - var)
        (charge_amount ?r - var ?t - var)
        (help_topic ?r - var ?t - var)
        (contact_us_topic ?r - var ?t - var)
        (contact_us_channel ?r - var ?t - var)
        (invoice_amount ?r - var ?t - var)
        (invoice_detail ?r - var ?t - var)
        (customer_given_name ?r - var ?t - var)
        (customer_family_name ?r - var ?t - var)
        (customer_email ?r - var ?t - var)
        (customer_phone ?r - var ?t - var)
        (has_type ?a - var ?t - var_type)
        (has_value ?a - var)
     )

(:action get_info_api
    :parameters (?in_var - var ?in_type - var_type)
    :precondition (and (has_type ?in_var ?in_type)
        (not (has_value ?in_var)))
    :effect (and (has_value ?in_var)))

(:action profit_loss_api
    :parameters (?in1 - var ?in2 - var ?out - var)
    :precondition (and (has_type ?in1 date) (has_value ?in1)
        (has_type ?in2 date) (has_value ?in2)
        (has_type ?out profit_loss_report) (not (has_value ?out)))
    :effect (and (report_start_date ?out ?in1)
        (report_end_date ?out ?in2) (has_value ?out)))

(:action expense_spend_api
    :parameters (?in1 - var ?in2 - var ?out - var)
    :precondition (and (has_type ?in1 date) (has_value ?in1)
        (has_type ?in2 date) (has_value ?in2)
        (has_type ?out expense_spend_report) (not (has_value ?out)))
    :effect (and (report_start_date ?out ?in1)
        (report_end_date ?out ?in2) (has_value ?out)))

(:action invoice_sales_api
    :parameters (?in1 - var ?in2 - var ?out - var)
    :precondition (and (has_type ?in1 date) (has_value ?in1)
        (has_type ?in2 date) (has_value ?in2)
        (has_type ?out invoice_sales_report) (not (has_value ?out)))
    :effect (and (report_start_date ?out ?in1)
        (report_end_date ?out ?in2) (has_value ?out)))

(:action charge_lookup_api
    :parameters (?in1 - var ?in2 - var ?out - var)
    :precondition (and (has_type ?in1 date) (has_value ?in1)
        (has_type ?in2 number) (has_value ?in2)
        (has_type ?out charge_lookup_report) (not (has_value ?out)))
    :effect (and (charge_date ?out ?in1) (charge_amount ?out ?in2)
        (has_value ?out)))

(:action help_api
    :parameters (?in1 - var ?out - var)
    :precondition (and (has_type ?in1 string) (has_value ?in1)
        (has_type ?out help) (not (has_value ?out)))
    :effect (and (help_topic ?out ?in1) (has_value ?out)))

(:action contact_us_api
    :parameters (?in1 - var ?in2 - var ?out - var)
    :precondition (and (has_type ?in1 contact_topic) (has_value ?in1)
        (has_type ?in2 contact_channel) (has_value ?in2)
        (has_type ?out contact) (not (has_value ?out)))
    :effect (and (contact_us_topic ?out ?in1)
        (contact_us_channel ?out ?in2) (has_value ?out)))

(:action create_invoice_api
    :parameters (?in1 - var ?in2 - var ?out - var)
    :precondition (and (has_type ?in1 number) (has_value ?in1)
        (has_type ?in2 string) (has_value ?in2) (has_type ?out invoice)
        (not (has_value ?out)))
    :effect (and (invoice_amount ?out ?in1) (invoice_detail ?out ?in2)
        (has_value ?out)))

(:action update_customer_api
    :parameters (?in1 - var ?out - var)
    :precondition (and 
        (has_type ?in1 customer_given_name) (has_value ?in1)
        (has_type ?in2 customer_family_name) (has_value ?in2)
        (has_type ?in3 customer_email) (has_value ?in3)
        (has_type ?in4 customer_phone) (has_value ?in4)
        (has_type ?out customer))
    :effect (and (customer_given_name ?out ?in1)
        (customer_family_name ?out ?in2) (customer_email ?out ?in3)
        (customer_phone ?out ?in4) (has_value ?out)))
)
\end{verbatim}
\end{tcolorbox}

\subsection{Domain Rules}
\label{appendix:domain-rules}
\begin{tcolorbox}[left=0pt,right=0pt,enhanced,breakable,size=fbox]
\begin{verbatim}
goal(X, profit_loss_report) :- _goal(X, goal_1, _).
start_date(X, Y1, date) :- goal(X, profit_loss_report),
    _report_period(X, (Y1, Y2)).
end_date(X, Y2, date) :- goal(X, profit_loss_report), 
    _report_period(X, (Y1, Y2)).

goal(X, expense_spend_report) :- _goal(X, goal_2, _).
start_date(X, Y1, date) :- goal(X, expense_spend_report), 
    _report_period(X, (Y1, Y2)).
end_date(X, Y2, date) :- goal(X, expense_spend_report),
    _report_period(X, (Y1, Y2)).

goal(X, invoices_sales_report) :- _goal(X, goal_3, _).
start_date(X, Y1, date) :- goal(X, invoices_sales_report),
    _report_period(X, (Y1, Y2)).
end_date(X, Y2, date) :- goal(X, invoices_sales_report),
    _report_period(X, (Y1, Y2)).

goal(X, charge_lookup) :- _goal(X, goal_4, _).
charge_date(X, Y, date) :- goal(X, charge_lookup),
    _date_of_charge(X, Y).
charge_amount(X, Y, number) :- goal(X, charge_lookup),
    _amount_of_charge(X, Y).

goal(X, helpgpt) :- _goal(X, goal_5).
help_topic(X, Y, string):- _help_topic(X, Y).

goal(X, contact_us) :- _goal(X, goal_6, _).
contact_topic(X, Y, fuzzy_string) :- goal(X, contact_us),
    _contact_topic(X, Y).
contact_channel(X, Y, fuzzy_string) :- goal(X, contact_us),
    _contact_channel(X, Y), Y == "video".
contact_channel(X, Y, fuzzy_string) :- goal(X, contact_us),
    _contact_channel(X, Y), Y == "chat".
contact_channel(X, Y, fuzzy_string) :- goal(X, contact_us),
    _contact_channel(X, Y), Y == "phone".

goal(X, advice) :- _goal(X, goal_7).

goal(X, create_invoice) :- _goal(X, goal_8, new_invoice).
invoice_amount(X, Y, number) :- goal(X, create_invoice),
    _invoice_amount(X, Y).
invoice_detail(X, Y, fuzzy_string) :- goal(X, create_invoice),
    _invoice_detail(X, Y).

goal(X, update_customer) :- _goal(X, goal_9, customer_profile).
customer_given_name(X, Y, string) :- goal(X, update_customer),
    _customer_given_name(X, Y).
customer_family_name(X, Y, string) :- goal(X, update_customer),
    _customer_family_name(X, Y).
customer_email(X, Y, string) :- goal(X, update_customer),
    _customer_email(X, Y).
customer_phone(X, Y, string) :- goal(X, update_customer),
    _customer_phone(X, Y).

error("end date must be after start date") :- start_date(X, D1, date),
    end_date(X, D2, date), false == @lte_dates(D1, D2).
...
\end{verbatim}
\end{tcolorbox}

\section{Query ASP to Task PDDL}
\subsection{Query ASP to Task PDDL Algorithm}
\label{appendix:mat-rep-to-task-pddl-algo}

\begin{tcolorbox}[left=0pt,right=0pt,breakable,enhanced,size=fbox]
\begin{algorithmic}    
    \State str\_objects $\gets$ ""
    \State str\_init $\gets$ ""
    \State str\_goal $\gets$ ""
    \State Initialize objects $\gets \{var\_type : \emptyset, var : \emptyset \}$. Initialize all\_vars $\gets \emptyset$.
    \For{each goal in goals}
        \State Add goal to objects[var\_type].
        \For{each var in goals[goal]}
            \State add goals[goal][var][type] to objects[var\_type]
            \State add var to all\_vars
        \EndFor
    \EndFor
    \For{each atom in the ASP model}
        \State goal\_var, goal\_name $\gets$ atom.arguments[0], atom.arguments[1]
        \If{atom.name == "goal"}
            \State Add goal\_var to objects[var]
            \State str\_init += "(has\_type " + goal\_var + " " + goal\_name + ")"            
            \For{each arg of goals[goal\_name]}
                \State arg\_var $\gets$ goal\_var+"\_"+arg
                \State arg\_type $\gets$ goals[goal\_name][arg]["type"]
                \State pred\_name $\gets$ goals[goal\_name][arg]["predicate"]
                \State Add arg\_var to $objects[var]$
                \State str\_init += "(has\_type " + arg\_var + " "  + arg\_type + ")"
                \State str\_goal += "(" + pred\_name + " " + goal\_var + " " + arg\_var + ")"
            \EndFor
        \ElsIf{atom.name in all\_vars}
            \State goal\_var $\gets$ atom.arguments[0].name
            \State  var $\gets$ goal\_var + "\_" + atom.name
            \State str\_goal += "(" + atom.name + " " + goal\_var + " " + var + ")"            
        \EndIf        
    \EndFor
    
    \State str\_objects += " ".join(objects[var]) + " - " + var
    \State str\_objects += " ".join(objects[var\_type]) + " - " + var\_type
\end{algorithmic}
\end{tcolorbox}

\subsection{Example Task PDDLs}
\label{appendix:example-task-pddl}

\begin{tcolorbox}[left=0pt,right=0pt,breakable,enhanced,size=fbox]
\begin{verbatim}
(define (problem example1)
    (:domain query-to-plan)
    (:objects 
        profit_loss_report - var_type
        x_end_date x_start_date x - var
    )
    (:init 
        (has_type x profit_loss_report)
        (has_type x_start_date date)
        (has_type x_end_date date)
    )
    (:goal (and 
        (report_start_date x x_start_date)
        (report_end_date x x_end_date)
    ))
)
\end{verbatim}    
\end{tcolorbox}

\begin{tcolorbox}[left=0pt,right=0pt,breakable,enhanced,size=fbox]
\begin{verbatim}
(define (problem example2)
    (:domain query-to-plan)
    (:objects 
        expense_spend_report - var_type
        x_end_date x_start_date x - var
    )
    (:init 
        (has_type x profit_loss_report)
        (has_type x_start_date date)
        (has_value x_start_date "01/01/2023")
        (has_type x_end_date date)
        (has_value x_end_date "03/31/2023")
    )
    (:goal (and 
        (report_start_date x x_start_date)
        (report_end_date x x_end_date)
    ))
)
\end{verbatim}    
\end{tcolorbox}

\begin{tcolorbox}[left=0pt,right=0pt,breakable,enhanced,size=fbox]
\begin{verbatim}
(define (problem example3)
    (:domain query-to-plan)
    (:objects 
        contact_us - var_type
        x_topic x_channel x - var
    )
    (:init 
        (has_type x contact_us)
        (has_type x_topic string)
        (has_type x_channel string)
        (has_value x_channel "chat")
    )
    (:goal (and 
        (contact_us_topic x x_topic)
        (contact_us_channel x x_channel)
    ))
)
\end{verbatim}    
\end{tcolorbox}

\begin{tcolorbox}[left=0pt,right=0pt,breakable,enhanced,size=fbox]
\begin{verbatim}
(define (problem example4)
    (:domain query-to-plan)
    (:objects 
        contact date contact_channel contact_topic
        profit_loss_report - var_type
        y x_end_date y_contact_channel x_start_date
        y_contact_topic x - var
    )
    (:init 
        (has_type y contact)
        (has_type y_contact_topic contact_topic)
        (has_type y_contact_channel contact_channel)
        (has_type x profit_loss_report)
        (has_type x_start_date date)
        (has_type x_end_date date)
        (has_value x_start_date)
        (value x_start_date "last quarter start")
        (has_value x_end_date)
        (value x_end_date "last quarter end")
        (has_value y_contact_channel)
        (value y_contact_channel "phone")
    )
    (:goal (and 
        (contact_us_topic y y_contact_topic)
        (contact_us_channel y y_contact_channel)
        (report_start_date x x_start_date)
        (report_end_date x x_end_date)
        (contact_channel y y_contact_channel)        
    ))
)
\end{verbatim}    
\end{tcolorbox}

\section{Evaluation}
\subsection{LLM prompt for dataset generation}
\label{appendix:dataset-generation-prompt}
Example prompt for generating user query and entities related to expense report
\begin{tcolorbox}[left=0pt,right=0pt,breakable,enhanced,size=fbox]
\begin{verbatim}
"Write 20 questions that use the variables below. These questions will 
be used to test entity extraction.
The variables are
startperiod: the start date for the period of the expense and spend
endperiod: the end date the user wants for the expense and spend ,

Your response should be in the format following these examples:
{""Question"": ""spending breakdown"",
""startperiod"": [],
""endperiod"":[]
}

{""Question"": ""what have i spent most on 2020"",
""startperiod"": 1/1/2020,
""endperiod"": 12/31/2021
}

{""Question"": ""top monthly expenses from april 1 to may 2023"",
""startperiod"": 04/1/2023,
""endperiod"": 05/31/2023
}

{""Question"": ""top spending categories from 1/1/24 to 2/1/24"",
""startperiod"": 1/1/24,
""endperiod"": 2/1/24
}
"
\end{verbatim}    
\end{tcolorbox}

\subsection{Samples from the dataset }
\label{appendix:dataset-samples}
\begin{table}[h]
\centering
\begin{tabularx}{\textwidth} { 
  >{\centering\arraybackslash}X
  >{\centering\arraybackslash}X
  >{\centering\arraybackslash}X
  >{\centering\arraybackslash}X
  >{\centering\arraybackslash}X
  >{\centering\arraybackslash}X }
 \toprule
 Query & gt\_API & gt\_entity1 & gt\_value1 & gt\_entity2 & gt\_value2 \\
 \midrule
 Q1 2023 P\&L review? & profit\_loss & startperiod & 1/1/23 & endperiod & 3/31/23 \\
 Why was I charged \$75? & charge\_lookup & dateofcharge & [] & amountofcharge & 75 \\
 \bottomrule
\end{tabularx}
\end{table}

\subsection{Baseline Prompt}
\label{appendix:baseline-prompt}
\begin{tcolorbox}[left=0pt,right=0pt,breakable,enhanced,size=fbox]
\begin{verbatim}
{"role": "system", "content": """
Only use the functions you have been provided with. Do not assume or 
hallucinate function parameters. If user has not provided, ask user for 
required parameters. 
Don't make assumptions about what values to plug into functions. 
Ask for clarification if a user request is ambiguous.
"""},
{"role": "user", "content": query}
\end{verbatim}    
\end{tcolorbox}
Here, functions are the APIs modelled as OpenAI function specifications and query refers to the user query of interest.  

\subsection{Evaluation Results}
\label{appendix:evaluation-results}

\begin{table}[h]
\centering
\begin{tabularx}{\textwidth} { 
  m{1.2in}
  m{0.2in}
  >{\centering\arraybackslash}X
  >{\centering\arraybackslash}X
  >{\centering\arraybackslash}X
  >{\centering\arraybackslash}X
  m{0.2in}
  >{\centering\arraybackslash}X
  >{\centering\arraybackslash}X
  >{\centering\arraybackslash}X
  >{\centering\arraybackslash}X }
 \toprule
 & \multicolumn{5}{c}{\textbf{Complete Queries}} & \multicolumn{5}{c}{\textbf{Incomplete Queries}} \\ 
 & \textbf{\#} & \multicolumn{2}{c}{\textbf{GPT-4}} & \multicolumn{2}{c}{\textbf{GPT-3.5}} & \textbf{\#} & \multicolumn{2}{c}{\textbf{GPT-4}} & \multicolumn{2}{c}{\textbf{GPT-3.5}}\\
 & & Baseline & Our Approach & Baseline & Our Approach & & Baseline & Our Approach & Baseline & Our Approach\\
 profit \& loss report &70 &22.86 &\textbf{98.57} & 97.14&\textbf{100} &2 &0 & \textbf{100} & \textbf{100}&\textbf{100}\\
 \midrule
 expense spend report &42 &30.95 &\textbf{100} &97.62 &\textbf{100} &0 &- & -&-&-\\
 \midrule
 invoice sales report &33 &63.64 &\textbf{90.91} &87.88 &\textbf{93.94} &12 &16.67 & \textbf{100}&66.67&\textbf{100}\\
\midrule
 charge lookup &33 &81.82 &\textbf{100} &\textbf{100} &96.97 &5 &40.00 & \textbf{100}&\textbf{100}&\textbf{100}\\
\midrule
 how-to help &60 &70.00 &\textbf{98.33} &68.33 &\textbf{90.00} & 0&- & -&-&-\\
\midrule
 contact us request &10 &40.00 &\textbf{100} &80.00 &\textbf{100} &47 &14.89 &\textbf{93.62}&44.68&\textbf{95.74}\\
\midrule
 financial advice &100 &81.00 &\textbf{94.90} &94.00 &\textbf{97.00} &0 &- &-&-&-\\
\midrule
 create invoice &40 &60.00 &\textbf{100} &\textbf{100} &\textbf{100} &20 &0 &\textbf{100}&\textbf{100}&\textbf{100}\\
\midrule
 update customer &3 &0 &\textbf{100} &100 &\textbf{100} &30 &6.67 &\textbf{100}&\textbf{100}&\textbf{100}\\
\bottomrule
\end{tabularx}
\label{tbl:single-plugin-accuracy-complete-queries-gpt-4}
\caption{\textbf{API orchestration success rate \%} of our approach compared with a baseline of end-to-end LLM based approach on \textbf{single goal queries}}
\end{table}

\begin{table}
\centering
\begin{tabularx}{\textwidth} { 
  m{1.2in}
  m{0.2in}
  >{\centering\arraybackslash}X
  >{\centering\arraybackslash}X
  >{\centering\arraybackslash}X
  >{\centering\arraybackslash}X
  m{0.2in}
  >{\centering\arraybackslash}X
  >{\centering\arraybackslash}X
  >{\centering\arraybackslash}X
  >{\centering\arraybackslash}X }
 \toprule
 & \multicolumn{5}{c}{\textbf{Complete Queries}} & \multicolumn{5}{c}{\textbf{Incomplete Queries}} \\ 
 & \textbf{\#} & \multicolumn{2}{c}{\textbf{GPT-4}} & \multicolumn{2}{c}{\textbf{GPT-3.5}} & \textbf{\#} & \multicolumn{2}{c}{\textbf{GPT-4}} & \multicolumn{2}{c}{\textbf{GPT-3.5}}\\
 & & Baseline & Our Approach & Baseline & Our Approach & & Baseline & Our Approach & Baseline & Our Approach\\
 \midrule
 profit \& loss report &70 &22.86 &\textbf{98.57} &81.43 &\textbf{100} &2 &0 &\textbf{100} &0 &\textbf{100}\\
 \midrule
 expense spend report &42 &23.81 &\textbf{100} & 90.48&\textbf{100} &0 &- &-&-&-\\
 \midrule
 invoice sales report &33 &54.55 &\textbf{96.97} &84.85 &\textbf{96.97} &12 &0 &\textbf{100}&0&\textbf{91.67}\\
\midrule
 charge lookup &33 &81.82 &\textbf{100} &\textbf{96.97} &93.94 &5 &40.00 &\textbf{100}&0&\textbf{100}\\
\midrule
 how-to help &60 &68.33 &\textbf{98.33} &68.33 &\textbf{95.00} &0 &- &-&-&-\\
\midrule
 contact us request &10 &40.00 &\textbf{100} &70.00 &\textbf{100} &47 &0 &\textbf{97.87}&0&\textbf{87.23}\\
\midrule
 financial advice &100 & 81.00&\textbf{99.00} &94.00 &\textbf{100} &0 & -&-&-&-\\
\midrule
 create invoice &40 &57.50 &\textbf{100} &\textbf{100} &\textbf{100} &20 &0 &\textbf{100}&0&\textbf{100}\\
\midrule
 update customer &3 &0 &\textbf{100} &\textbf{100} &\textbf{100} &30 &6.67 &\textbf{100}&6.67&\textbf{100}\\
\bottomrule
\end{tabularx}
\label{tbl:single-plugin-accuracy-incomplete-queries-gpt-4}
\caption{\textbf{Entity extraction success rate \%} of our approach compared with a baseline of end-to-end LLM based approach on \textbf{single goal queries}}
\end{table}

\begin{table}
\centering
\begin{tabularx}{\textwidth} { 
  m{1.2in}
  m{0.2in}
  >{\centering\arraybackslash}X
  >{\centering\arraybackslash}X
  >{\centering\arraybackslash}X
  >{\centering\arraybackslash}X
  m{0.2in}
  >{\centering\arraybackslash}X
  >{\centering\arraybackslash}X
  >{\centering\arraybackslash}X
  >{\centering\arraybackslash}X }
 \toprule
 & \multicolumn{5}{c}{\textbf{Complete Queries}} & \multicolumn{5}{c}{\textbf{Incomplete Queries}} \\ 
 & \textbf{\#} & \multicolumn{2}{c}{\textbf{GPT-4}} & \multicolumn{2}{c}{\textbf{GPT-3.5}} & \textbf{\#} & \multicolumn{2}{c}{\textbf{GPT-4}} & \multicolumn{2}{c}{\textbf{GPT-3.5}}\\ 
 & & Base-line & Our Approach & Base-line & Our Approach & & Base-line & Our Approach & Base-line & Our Approach\\
 \midrule
 2 APIs w/o dataflow & 15 & 0 & \textbf{100}& 0 & \textbf{100} & 10 & 0 &\textbf{100} & 0 & \textbf{100}\\
 \midrule
 2 APIs with dataflow & 20 & 0 &\textbf{100} & 0 & \textbf{80} & 10 & 0 & \textbf{80} & 0 & \textbf{80}\\
\midrule
 3 APIs with dataflow & 4 & 0 &\textbf{100} & 0 & \textbf{75} & 16 & 0 & \textbf{100} & 0 & \textbf{81.25}\\
\bottomrule
\end{tabularx}
\label{tbl:multi-plugin-accuracy-complete-queries-gpt-4}
\caption{\textbf{API orchestration success rate \%} of our approach compared with a baseline of end-to-end LLM based approach on \textbf{multi goal complete queries}}
\end{table}

\begin{table}
\centering
\begin{tabularx}{\textwidth} { 
  m{1.2in}
  m{0.2in}
  >{\centering\arraybackslash}X
  >{\centering\arraybackslash}X
  >{\centering\arraybackslash}X
  >{\centering\arraybackslash}X
  m{0.2in}
  >{\centering\arraybackslash}X
  >{\centering\arraybackslash}X
  >{\centering\arraybackslash}X
  >{\centering\arraybackslash}X }
 \toprule
 & \multicolumn{5}{c}{\textbf{Complete Queries}} & \multicolumn{5}{c}{\textbf{Incomplete Queries}} \\ 
 & \textbf{\#} & \multicolumn{2}{c}{\textbf{GPT-4}} & \multicolumn{2}{c}{\textbf{GPT-3.5}} & \textbf{\#} & \multicolumn{2}{c}{\textbf{GPT-4}} & \multicolumn{2}{c}{\textbf{GPT-3.5}}\\ 
 & & Base-line & Our Approach & Base-line & Our Approach & & Base-line & Our Approach & Base-line & Our Approach\\
 \midrule
 2 APIs w/o dataflow & 15 & 0 & \textbf{100} & 0 & \textbf{100} & 10 & 0 & \textbf{100} & 0 & \textbf{100}\\
 \midrule
 2 APIs with dataflow & 20 & 0 & \textbf{90.00} & 0 & \textbf{70} & 10 & 0 & \textbf{90} & 0 & \textbf{60}\\
\midrule
 3 APIs with dataflow & 4 & 0 & \textbf{100} & 0 & \textbf{75.00} & 16 & 0 & \textbf{75} & 0 & \textbf{62.50}\\
\bottomrule
\end{tabularx}
\label{tbl:multi-plugin-accuracy-incomplete-queries-gpt-4}
\caption{\textbf{Entity extraction success rate \%} of our approach compared with a baseline of end-to-end LLM based approach on \textbf{multi goal queries}}
\end{table}

\end{document}